\documentclass[conference]{IEEEtran}
\IEEEoverridecommandlockouts
\usepackage{cite}
\usepackage{amsmath,amssymb,amsfonts}
\usepackage{algorithmic}
\usepackage{graphicx}
\usepackage{textcomp}
\usepackage{subcaption}
\usepackage{multirow}
\usepackage{comment}
\usepackage[usenames,dvipsnames]{xcolor}
\usepackage[font=small,skip=2pt]{caption}
\usepackage{xcolor}
\usepackage{hyperref}
\def\BibTeX{{\rm B\kern-.05em{\sc i\kern-.025em b}\kern-.08em
    T\kern-.1667em\lower.7ex\hbox{E}\kern-.125emX}}

\usepackage[pages=some,placement=top, scale=1,
hshift=0,
vshift=-20,]{background}
\usepackage{lipsum}
\backgroundsetup{contents=This paper has been accepted for publication at 2022 IEEE International Conference on
Emerging Technologies and Factory Automation (ETFA),firstpage=true, angle=0,color=gray, opacity=1}

\begin{document}
\title{COVERED, CollabOratiVE Robot Environment Dataset for 3D Semantic segmentation\\
\thanks{$^{5}$Corresponding author}
\thanks{* Indicates equal contributions from authors.}

}

\author{\IEEEauthorblockN{1\textsuperscript{st} Charith Munasinghe$^{*}$}\IEEEauthorblockA{\textit{Institute of Mechatronics Systems} \\
\textit{Zurich University of Applied Sciences} \\
Winterthur, Switzerland \\
\textit{mung@zhaw.ch}}
\and
\IEEEauthorblockN{2\textsuperscript{nd} Fatemeh Mohammadi Amin$^{*}$}
\IEEEauthorblockA{\textit{Institute of Mechatronics Systems} \\
\textit{Zurich University of Applied Sciences}\\
Winterthur, Switzerland \\
\textit{mohm@zhaw.ch}}
\and
\IEEEauthorblockN{3\textsuperscript{rd} Davide Scaramuzza}
\IEEEauthorblockA{\textit{Robotics and Perception Group} \\
\textit{University of Zurich} \\
Zurich, Switzerland \\
\textit{sdavide@ifi.uzh.ch}}
\and
\IEEEauthorblockN{4\textsuperscript{th} Hans Wernher van de Venn$^{5}$ }
\IEEEauthorblockA{\textit{Institute of Mechatronics Systems} \\
\textit{Zurich University of Applied Sciences (ZHAW)}\\
Winterthur, Switzerland \\
\textit{vhns@zhaw.ch}
}
}

\maketitle

\begin{abstract}
Safe human-robot collaboration (HRC) has recently gained a lot of interest with the emerging Industry 5.0 paradigm. Conventional robots are being replaced with more intelligent and flexible collaborative robots (cobots). Safe and efficient collaboration between cobots and humans largely relies on the cobot's comprehensive semantic understanding of the dynamic surrounding of industrial environments.
Despite the importance of semantic understanding for such applications, 3D semantic segmentation of collaborative robot workspaces lacks sufficient research and dedicated datasets. The performance limitation caused by insufficient datasets is called 'data hunger' problem. To overcome this current limitation, this work develops a new dataset specifically designed for this use case, named "COVERED", which includes point-wise annotated point clouds of a robotic cell. Lastly, we also provide a benchmark of current state-of-the-art (SOTA) algorithm performance on the dataset and demonstrate a real-time semantic segmentation of a collaborative robot workspace using a multi-LiDAR system. The promising results from using the trained Deep Networks on a real-time dynamically changing situation shows that we are on the right track. 
Our perception pipeline achieves 20Hz throughput with a prediction point accuracy of $>$96\% and $>$92\% mean intersection over union (mIOU) while maintaining an 8Hz throughput. 

\end{abstract}


\section{Introduction}
Leveraging industry 5.0 concepts, robotic research has opened up numerous possibilities for flexible and intelligent ways of automation and collaboration between humans and robots. In fact, cobots are increasingly being used for flexible task accomplishment instead of traditional industrial robots~\cite{olender2019cobots} and can work in the same workspace as humans~\cite{vicentini2021collaborative}.

\begin{figure}[t]
    \centering
    \begin{subfigure}[t]{0.51\columnwidth}
        \centering
        \frame{\includegraphics[width=1\textwidth]{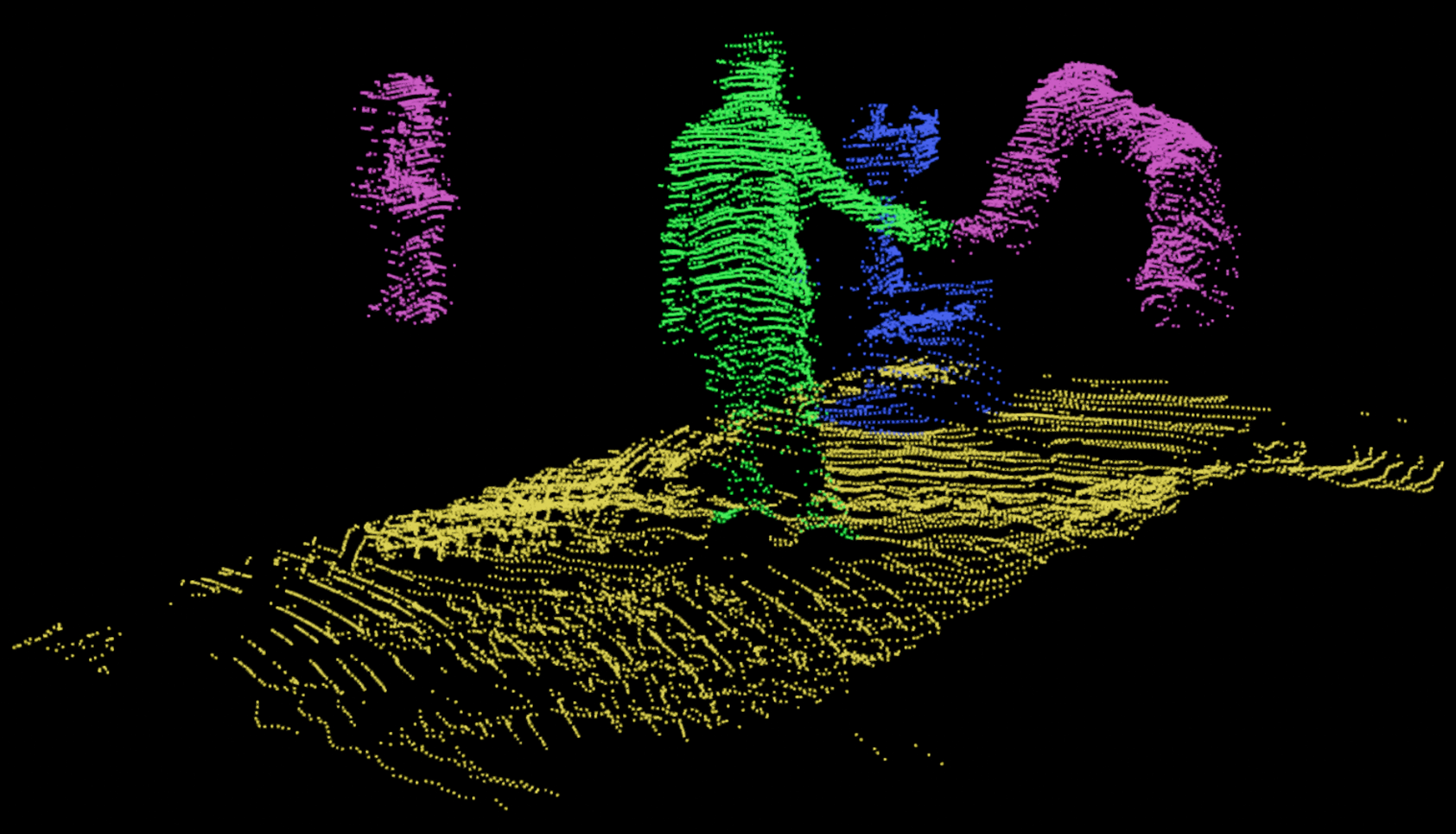}}
        \caption{Semantically-segmented Point cloud}
    \end{subfigure}%
    \begin{subfigure}[t]{0.47\columnwidth}
        \centering
        \frame{\includegraphics[width=1\textwidth]{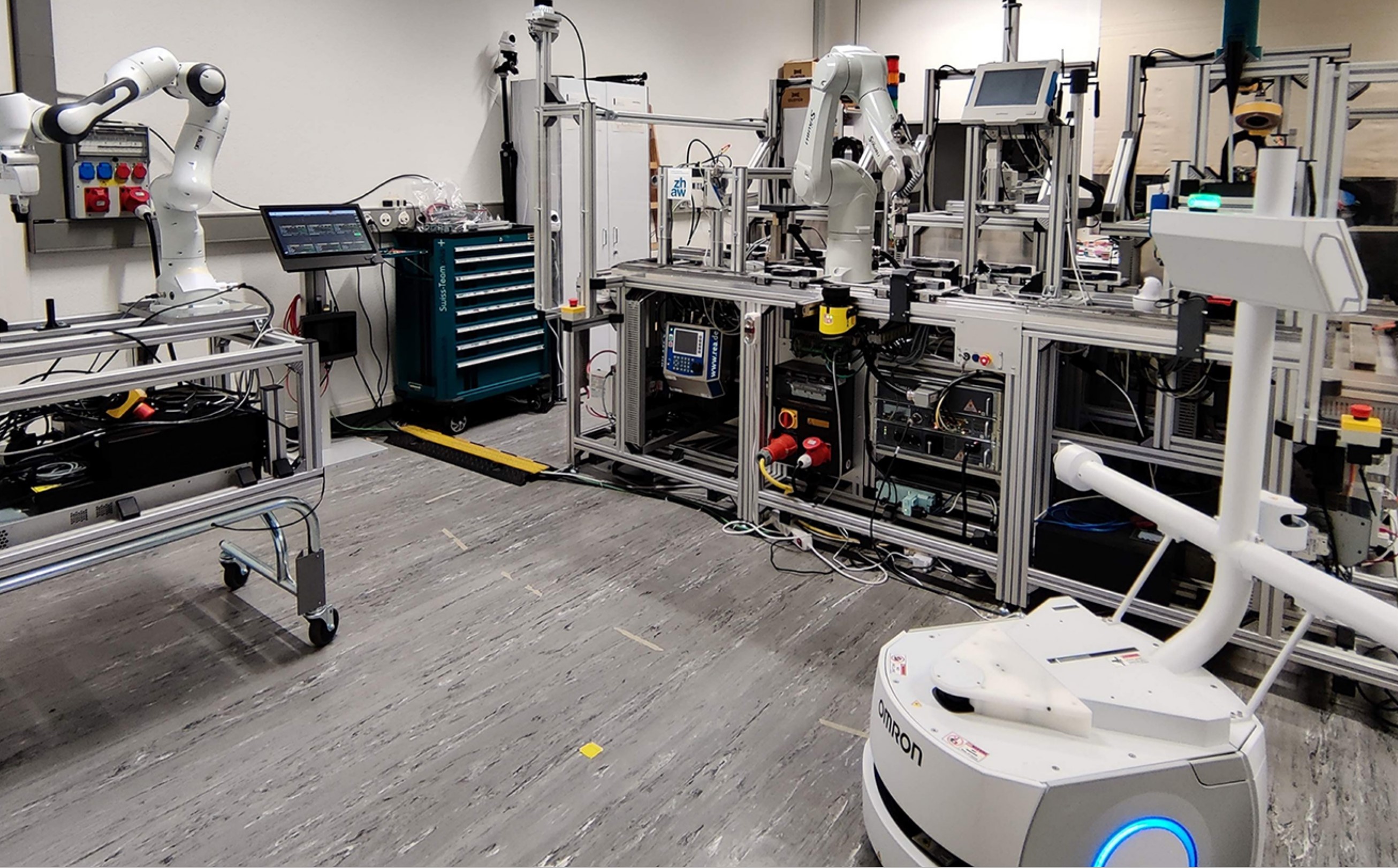}}
        \caption{Dynamic Collaborative area}
    \end{subfigure}
    \caption{Collaborative robotic workspace at IMS, ZHAW. }
    \vspace{-6mm}
\label{workspace}
\end{figure}

Therefore, cobots need to be significantly more intelligent than their conventional counterparts to be able to react to natural human inputs and dynamically changing environments in such a way that ensures smooth, safe, and productive workflows. Thus, sensing, perceiving, and understanding the environment in comprehensive detail is crucial and the artificial intelligence (AI) algorithms used should be able to anticipate and cope with different situations occurring in industrial environments~\cite{hamon2020robustness}. Semantic segmentation, which separates data of a given modality into semantically meaningful subsets, is fundamental to scene understanding~\cite{rangnet2019}. In the case of 3D point clouds, labeling each point with a predefined class allows to detect and distinguish objects precisely~\cite{guo2018review}.

Over the past decade, 3D semantic segmentation has developed rapidly as a field of research in robotics, especially in autonomous driving~\cite{fernandes2021point}.
For 3D semantic segmentation tasks, 3D LiDAR data with point-wise annotation are required, where S3DIS \cite{s3dis}, Semantic3D \cite{semantic3d}, and SemanticKITTI \cite{behley2019semantickitti} are among the most popular datasets for general applications. Due to the annotation difficulties, the publicly available datasets for 3D semantic segmentation are very limited in both data size and diversity compared to image datasets. 

\begin{table*}[h]
    \renewcommand{\arraystretch}{1}
    \caption{3D Lidar datasets}
    \label{tab1}
    \begin{center}
        \resizebox{0.8\textwidth}{!}{
        \begin{tabular}{|c|c|c|c|c|c|c|c|}
            \hline
            \textbf{dataType} & \textbf{\textit{dataset}}& \textbf{\textit{frames}}& \textbf{\textit{points}} & \textbf{\textit{classes}}& \textbf{\textit{Scene}} & \textbf{\textit{year}} & \textbf{objects} \\
            \hline\hline
            \multirow{3}{*}{Static} & S3DIS \cite{s3dis} & 5 & 215M & 12 & Indoor& 2017 & Static \\
            & Semantic3D \cite{semantic3d}& 30 & 4009M & 8 & Outdoor & 2017 & Static \\
            & Paris-Lille-3D \cite{paris2017} & 3 & 143M  & 50  & Outdoor & 2018 & Static \\ 
            \hline
            \multirow{4}{*}{Sequential} & SemanticKITTI \cite{behley2019semantickitti} & 20351  & 4549M  & 28 & Outdoor   & 2019 & D-S obj \\ & DALES \cite{Dales}  & 40 & 505M & 8                     & Outdoor & 2020& D-S obj \\ & SemanticPOSS \cite{Semanticposs} & 2988 &216M  & 14 & Outdoor  & 2020 & D-S obj \\ & KITTI-360 \cite{kitti360} & 100K & 18B & 19 & Outdoor& 2021 & D-S obj \\

            \hline
            \multirow{1}{*}{Static} & COVERED(ours) & 218 & 48M & 6 & Industrial Env & 2022  & D-S obj \\ 
            \hline
        \end{tabular}}
    \end{center}
    \vspace{-4mm}
\end{table*}

There is also an inadequacy of research focusing on semantic understanding in HRC applications. The majority of HRC research focuses on image-based data like RGB and RGBD, which contain occlusion problems and lack the 3D information that is critical for determining the accurate location of objects (such as humans and robots) for ensuring \textbf{human safety} during collaboration with robots~\cite{mohammadi2020mixed}. The lack of precise perception of the dynamic environment may result in fatal physical injuries to humans in the worst case~\cite{Jiang1987ACA}. Therefore, industrial robot cells are usually designed as fenced work areas, which human can not enter during operation, to ensure rigid safety standards.~\cite{9196924}. In contrast for dynamic safety, AI powered robots must be trained with appropriate datasets before they can execute AI algorithms in a real world application. These datasets must be carefully selected to provide the correct training data for every use case to not limit the performance of the system. The performance limitation caused by a lack of training data is called \textbf{data hunger effect} \cite{Hungry} which especially is a major obstacle in 3D semantic segmentation research of HRC applications. Table \ref{tab1} shows some of these datasets and their characteristics which illustrate better the data hunger for HRC. The static and sequential data type indicates that the data is captured from a fix or moving view point respectively. While some of these static datasets like Semantic3D contain no moving objects such as people, our dataset includes both dynamic and static objects (D-S obj). 

In this paper, we address the problems of lack of dataset, occlusion, and perceiving the industrial environment by developing an industrial dataset and demonstrate a multi-LiDAR 3D semantic segmentation system in a real industrial human-robot collaboration scenario. We further intend to use the dataset in such applications like semantic segmentation, completion networks and occlusion problems in industrial environments. The main contributions we make are as follows:
\begin{itemize}
    \item To the best of our knowledge, we present the first point-wise annotated dataset from a collaborative robotic workspace that includes multiple practical scenarios.
    \item We used the multi-LiDAR system to partly solve the occlusion problem and have a better distribution and resolution in our dataset. We evaluate the dataset using two SOTA deep learning models for 3D semantic segmentation of point clouds.
    \item We demonstrate a software stack that employs the above deep learning models for real-time semantic segmentation and explore the validity of the output for using it in high-level HRC applications.
\end{itemize}

\section{The dataset}

\subsection{Collaborative Workspace}
As shown in Fig \ref{Scenarios}, the collaborative workspace is a compact space with static and multiple dynamic objects including humans, cobots, and AGVs. The cobot has the task of assembling a customized pen from parts arriving in a conveyor carrier and an automated guided vehicle (AGV) moves the second cobot to the assembly station, where the main cobot is working, for supporting the task. After completion of the pen assembly, the cobot hands over the completed product to a human operator for inspection. The human operator controls the production and intervenes to instruct or correct the cobots when needed, whereas the AGV moves around in dynamically planned paths.
Considering the number of objects in this confined space, the environment poses many challenges. Occlusions are common because moving objects obscure the view to other items in different ways. Different reflection factors, shapes, and sizes of objects intensify the challenges in perception. To overcome some of these challenges, multiple LiDAR sensors are strategically positioned to capture the environment in high detail and to avoid full occlusions of objects.

\subsection{Preprocessing and Data Collection}
The data was collected using four Ouster OS0-128 LiDAR sensors and a host computer connected to a dedicated network to provide the required quality of service (QoS). 
As part of the initialization phase, the sensors are time-synchronized so the combined point clouds can be created from all sensors at the same timestamp.
The raw data needs to be filtered, registered, and aggregated to be used for machine learning and other systems. 

\subsection{COVERED dataset}

Data is captured at 20Hz with 1024x128 resolution which results in approximately 60,000 points per point cloud for each LiDAR sensor after filtering and trimming. In order to exclude redundant data and to easily annotate the unique configuration and scene, instead of using the 20Hz sample rate the dataset was annotated on a sample rate of 1Hz.
Each point cloud is manually annotated with six classes stated above, using a visual tool\footnote{\href{https://github.com/Hitachi\%2DAutomotive\%2DAnd\%2DIndustry\%2DLab/semantic-segmentation-editor}{Semantic Segmentation Editor}}. The dataset is available for public at the GitHub repository\footnote{\href{https://github.com/Fatemeh-MA/COVERED-A-dataset-for-3D-Semantic-segmentation.git}{COVERED Dataset GitHub Link}}. This repository includes 218 point-wise annotated point clouds in *.pcd format as well as *.npy format for efficient processing by machine learning tools. 

The points are annotated featuring six classes: Robots, Human, AGV, Floor, Wall, and Unspecified. The additional class "Unspecified" includes all other types of objects which are not of direct interest for the applications of this work. The average point density for these classes are like Robots: 1800 points, Human: 2800, AGV: 1200 ,Floor: 10000 and Wall: 13400 points in a 24 $m^2$ area. We intend to provide an extended dataset in the future with more data and classes.
\begin{figure*}[h]
    \centering
    \begin{subfigure}[t]{0.51\columnwidth}
        \centering
        \frame{\includegraphics[height=4cm]{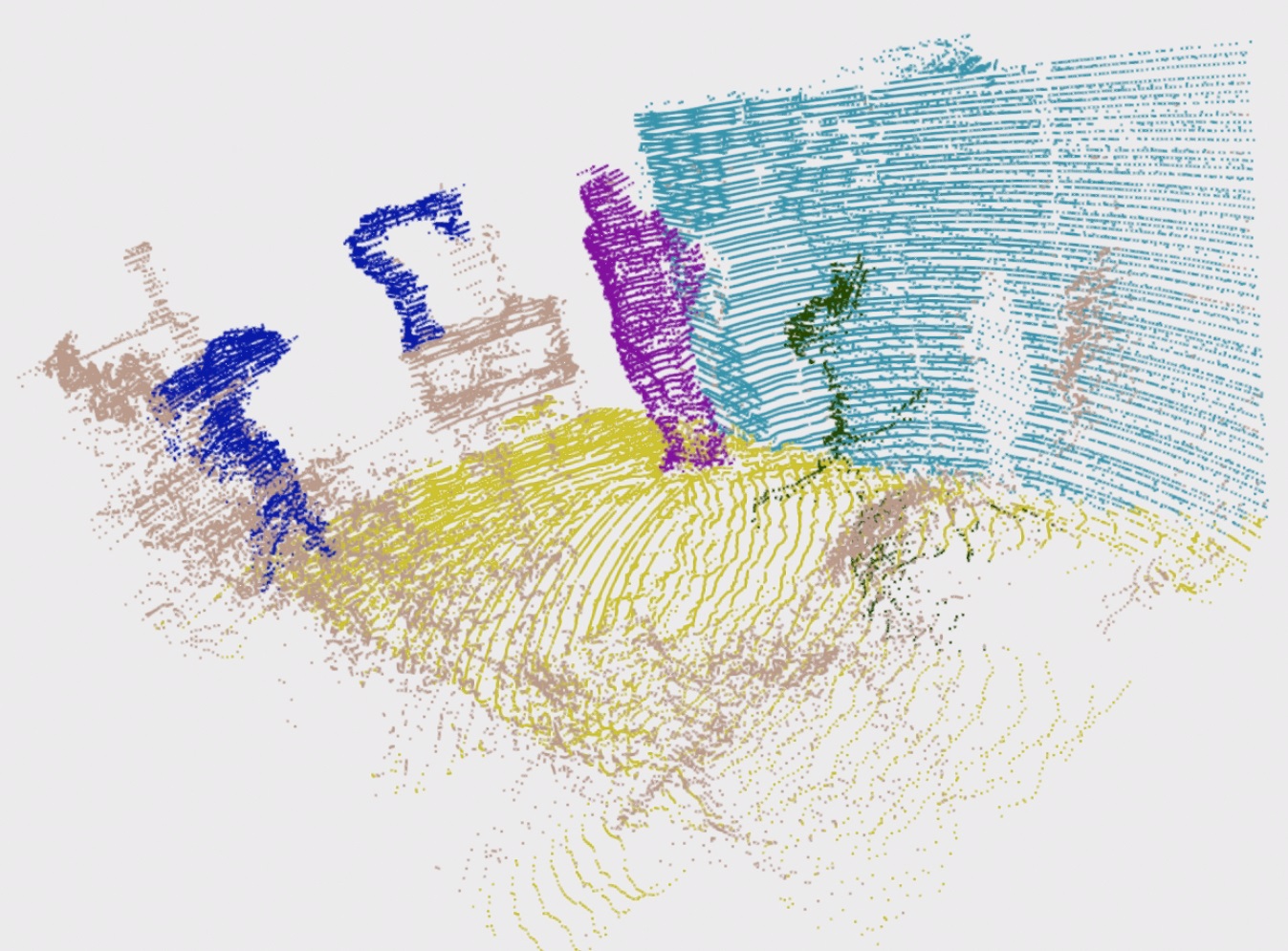}}
        \caption{Scenario 1}
    \end{subfigure}%
    \begin{subfigure}[t]{0.52\columnwidth}
        \centering
        \frame{\includegraphics[height=4cm]{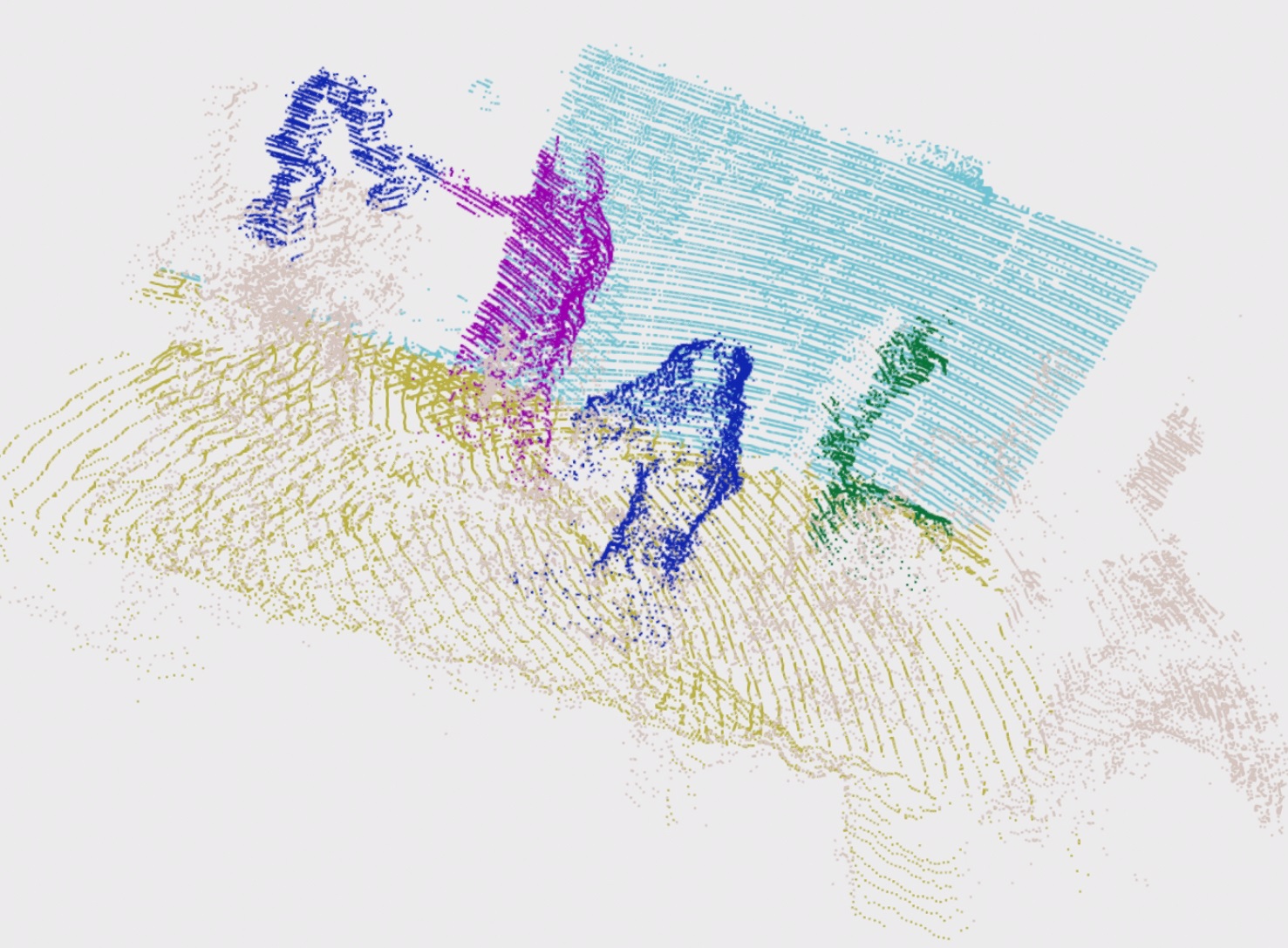}}
        \caption{Scenario 2}
    \end{subfigure}
    \begin{subfigure}[t]{0.44\columnwidth}
        \centering
        \frame{\includegraphics[height=4cm]{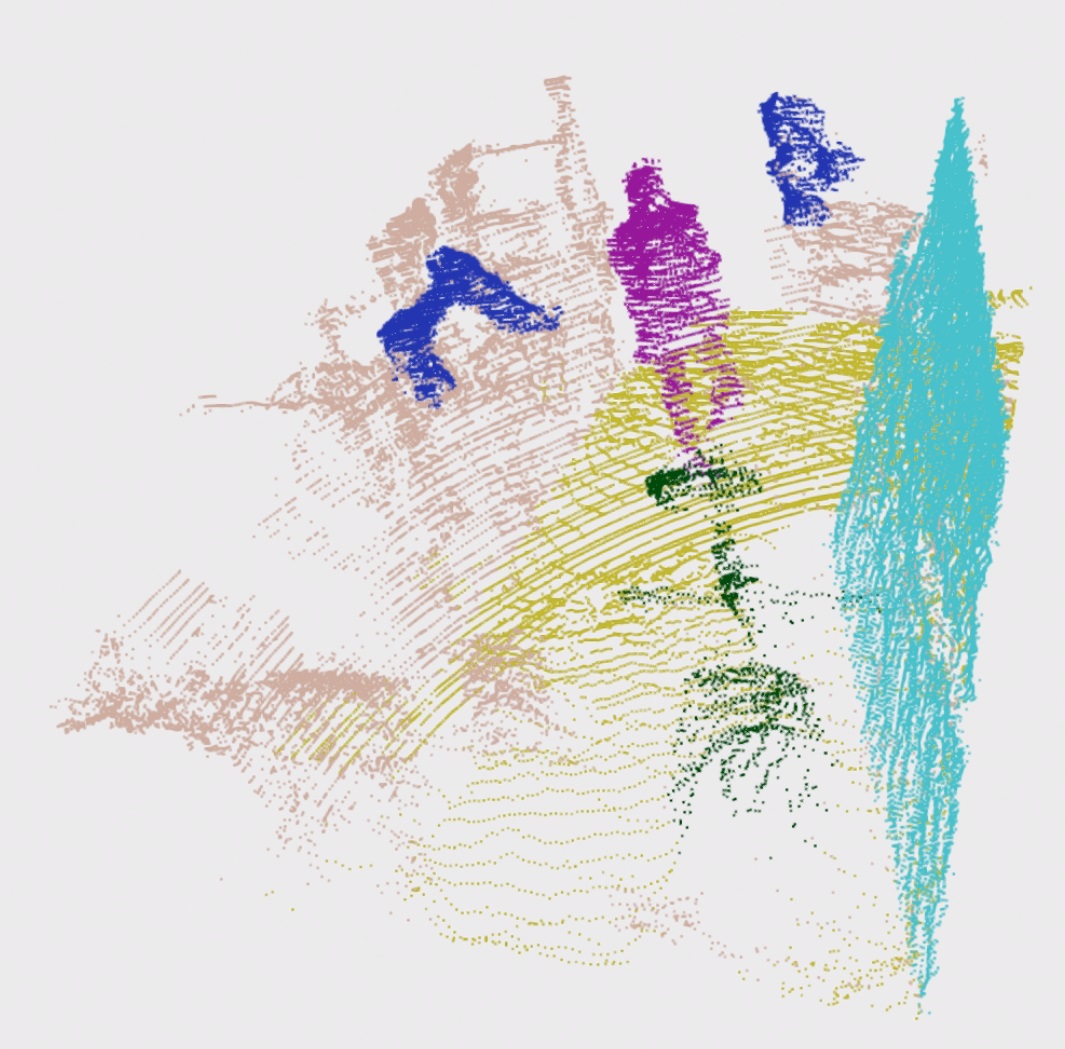}}
        \caption{Scenario 3}
    \end{subfigure}
    \begin{subfigure}[t]{0.48\columnwidth}
        \centering
        \frame{\includegraphics[height=4cm]{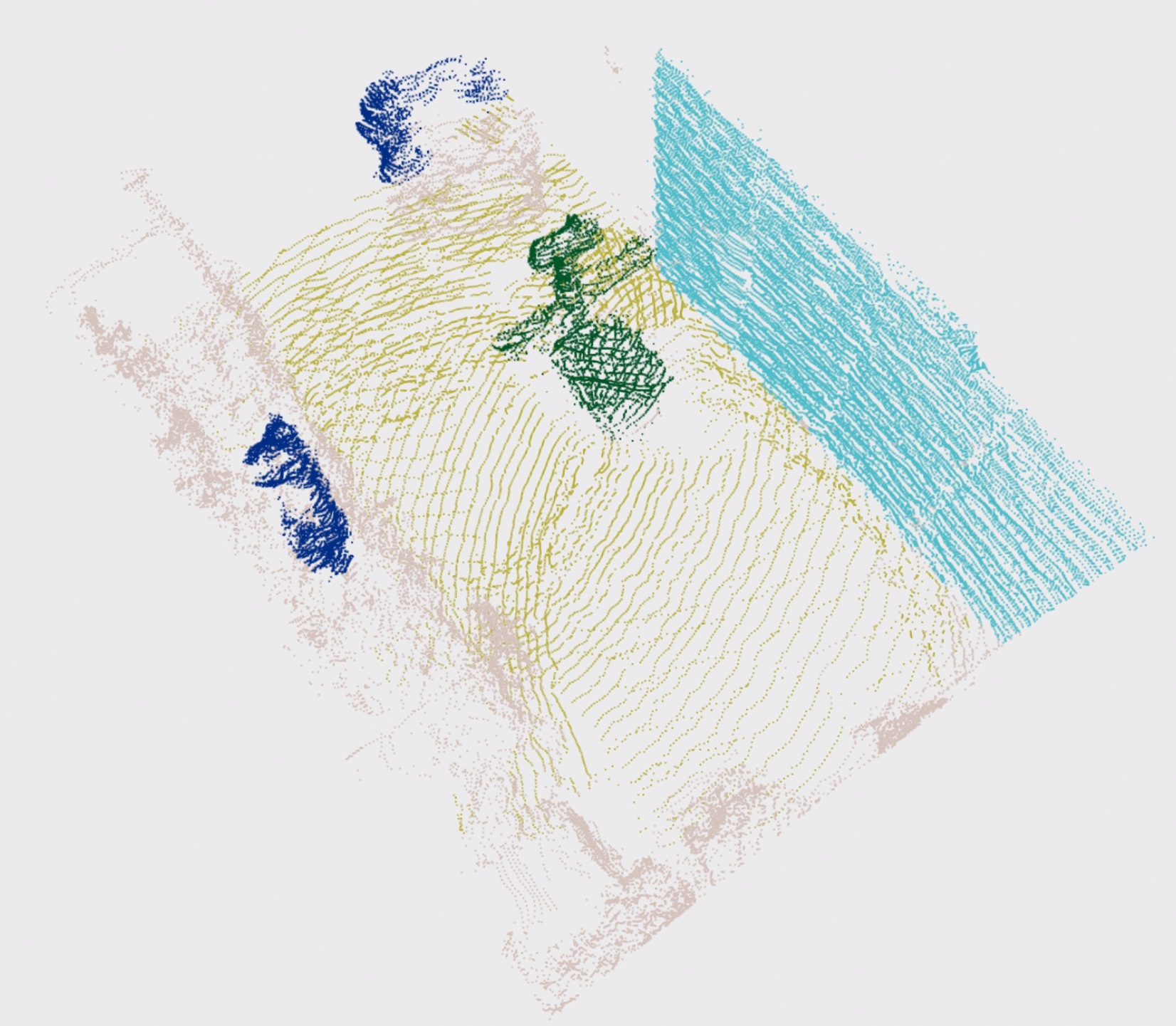}}
        \caption{Scenario 4}
    \end{subfigure}
    \caption{ Defined Scenarios for COVERED }
    \vspace{-5mm}
\label{Scenarios}
\end{figure*}

\subsection{Scenarios}
The dataset covers multiple practical and common scenarios (Fig \ref{Scenarios}) in the collaborative robotic workspace as follows:

\begin{enumerate}
    \item Two cobots are carrying out pre-programmed tasks. An operator observes the work and interacts with Human Machine Interface (HMI) and an AGV moves around.
    \item The operator interacts with one cobot to resume from an error state and with an other to receive the assembled product from robot gripper.
    \item The cobots are in usual operation and an operator passes by without interaction and collects completed products.
    \item Cobots and AGV are working autonomously without any operator presence or intervention.
\end{enumerate}

\section{Experimental Results}
\subsection{Evalution Metrics}\label{AA}
We follow the evaluation metrics of similar benchmarks like \cite{Dales} and use the mean IoU as our main metric. The IoU formula per class can be calculated by
\begin{equation}
IOU{_i}=\frac{c_{ii}} {c_{ii}+ \sum_{j \neq i} c_{ij} }\label{eq}
\end{equation}
we simply calculate the mean IoU of all six categories. As the second metric we calculated the overall accuracy (OA) as follows:
\begin{equation}
OA =\frac{\sum_{i=1}^{N} c_{ii}} {\sum_{j=1}^{N}\sum_{k=1}^{N} c_{jk}}\label{eq}
\end{equation}

Furthermore, many current studies assess model performance on a "closed set," assuming the testing set follows the same distribution as the training set.
Nevertheless, real-world applications are "open set" problems which require deep models to deal with new scenarios and scenes, and will always be data hungry in new scenes. Accordingly, another important evaluation for our system is the real-time testing which is one of the biggest achievements for this work and shows the robustness in the network performance on the dataset. It proves that our dataset has a very good distribution. A video from real-time testing is available under this \href{https://www.youtube.com/watch?v=UbHgE9W7144&t=28s}{YouTube link}.

\subsection{Algorithm Performance} 
 Semantic understanding begins with semantically segmenting the environment of interest. 3D semantic segmentation is often a supervised learning task that requires a point-wise annotated dataset of the environment. We selected two benchmark algorithms based on their strong performance on the mentioned datasets to evaluate their performance on our dataset. KPConv~\cite{thomas2019kpconv} and RandLA-Net \cite{hu2020randla} were selected as the best candidates to evaluate our dataset. Both models were trained and tested using the same train-, validation-, and test splits from the "COVERED" dataset with multi-fold cross-validation and examined by accuracy, OA, and mIOU.

In order to find the optimal hyper-parameters and model configurations, multiple tests were carried out. 
After achieving relatively high training performance, the model parameters were fixed and re-validated using the test split in the offline version. 
Table \ref{tab:testresults} shows the overall performance of the two models on the test data. Both models show more than 96\% accuracy and 92\% mIOU. This high accuracy was obtained due to the fact, that the dataset was able to properly describe the problem space of the application and both models were complex enough to describe the decision boundaries of the problem. This high accuracy was also clearly observed when visually inspecting the real-time predictions later.

\begin{table}[h]
    \centering
    \caption{Overall test accuracy and test mIOU for two models}
    \label{tab:testresults}
        \begin{tabular}{|c|c|c|}
        \hline
        \bfseries Metric           & \bfseries KPConv   & \bfseries RandLA-Net \\ \hline
        \bfseries Overall Accuracy & 0.976              & 0.960     \\ \hline
        \bfseries Overall IoU      & 0.946              & 0.927    \\ \hline
        \end{tabular}%
\end{table}

Table \ref{tab:classaccuracy} indicates the per-class accuracy each model obtained using the test split (30 percent) of the dataset. It was evident that both models were performing very well for most classes, but KPConve has shown a slightly better performance in detecting human and almost similar for robot which is of interest to us. 

\begin{table}[h]
\renewcommand{\arraystretch}{1.3}
    \centering
    \caption{Class accuracy and mIOU of models}
    \label{tab:classaccuracy}
        \resizebox{0.48\textwidth}{!}{%
        \begin{tabular}{|c|c|c|c|c|c|c|c|}
            \hline
            &        & Unspecified & Floor & Wall  & Robot & Human & AGV   \\ \hline
            \multirow{2}{*}{\bfseries Accuracy} & RandLA-Net  & 0.972  & 0.985 & 0.981 & 0.975 & 0.866 & 0.981 \\ \cline{2-8}
                                        & KPConv & 0.971       & 0.977 & 0.994 & 0.944 & 0.990 & 0.983 \\ \hline
            \multirow{2}{*}{\bfseries mIoU}       & RandLA-Net    & 0.972       & 0.929 & 0.961 & 0.919 & 0.786 & 0.949 \\ \cline{2-8} 
                                        & KPConv        & 0.962       & 0.930 & 0.963 & 0.916 & 0.958 & 0.951 \\ \hline
        \end{tabular}%
        }
\end{table}

In real-time testing, we also observed that KPConv performed better in defining the segmentation boundaries, especially when humans and objects are in proximity to each other. Considering the importance of safety to industrial scenarios, this is a huge advantage.

\section{Discussion, Conclusion and Outlook}

Despite the remarkable success of semantic segmentation techniques on the reviewed datasets, there is still a long way to go for robots to be able to perceive their surroundings in the same way humans do. On the other hand, since the annotation of real datasets is labor intensive, the generation of these datasets is very expensive, and to the best of our knowledge, there is no relevant 3D LiDAR dataset for industrial environments up to now. To fill this gap, we introduce COVERED, a CollabOratiVE Robot Environment dataset. As already mentioned, most known datasets focus on autonomous driving and static environments and only reflect a very small amount of real scenes, while our dataset covers a dynamic environment including humans, robots and 4 other distinguishable objects.

Despite some limitations, our dataset is quite sufficient for the first attempt at segmenting industrial environments. However, for a more accurate classification, especially in the close collaboration between humans and robots, it is necessary to distinguish between different robots and have extremely accurate real-time segmentation to ensure human safety. 
To this end, we are planning to complete the dataset, in both, class types and different scenes and scenarios.
Another important matter for analyzing the existing datasets is the statistics of point clouds.
A statistical analysis of the point number distribution of people and vehicle instances per-scene in SemanticKITTI and SemanticPOSS shows that more than half of instances contain fewer than 120 points, which does not contribute significantly to the training of models \cite{Hungry} and are difficult to recognize and distinguish even for humans; with more points, the features tend to be clearer to extract. Therefore, it is reasonable to use the point number as a measurement of instance quality.

To address this issue, we use a multi-LiDAR sensor and achieved a high point density.
Taking all these factors into account, robotics and autonomous driving in complex real-world scenarios may always suffer from data hunger\cite{Hungry}.
Therefore, in training and handling of rare/unseen objects, it is important to develop methods that do not rely on finely annotated data;
However, it is just as important as completing the datasets, especially for dynamic objects.
We also analyzed the dataset with two SOTA deep learning models and achieved excellent results in 3D semantic segmentation. Unfortunately, the results from benchmark datasets for other applications are not directly comparable to ours. However, our real-time perception and prediction pipeline that can directly be applied to industrial setups has shown amazing results on semantic segmentation, even for scenarios that are not in the training dataset (e.g., more humans, different robot, etc.). 
Thus, we believe, our dataset represents the problem space very well for this application and can be considered as a benchmark dataset for future research in similar applications. It will allow the research community to develop new algorithms based on it.

In the future, we plan to release an even larger dataset from our collaborative robot workspace with more scenarios and classes. In addition, we plan to improve the real-time performance of the pipelines and develop deep learning algorithm to keep up with the SOTA.

\section*{Acknowledgment}
We  gratefully  acknowledge  Phillip Steven Luchsinger and Alexander Wyss at IMS, ZHAW for their support in annotating the dataset. This work was supported by DIZH (Digitalization Initiative of the Zurich Higher education Institutions) funding.

\IEEEtriggeratref{22}
\bibliographystyle{ieeetr}
\bibliography{biblography.bib}

\begin{thebibliography}{10}

\bibitem{olender2019cobots}
M.~Olender and W.~Banas, ``Cobots--future in production,'' {\em International
  Journal of Modern Manufacturing Technologies, Special Issue, XI}, vol.~3,
  pp.~103--109, 2019.

\bibitem{vicentini2021collaborative}
F.~Vicentini, ``Collaborative robotics: a survey,'' {\em Journal of Mechanical
  Design}, vol.~143, no.~4, p.~040802, 2021.

\bibitem{hamon2020robustness}
R.~Hamon, H.~Junklewitz, and I.~Sanchez, ``Robustness and explainability of
  artificial intelligence,'' {\em Publications Office of the European Union},
  2020.

\bibitem{rangnet2019}
A.~Milioto, I.~Vizzo, J.~Behley, and C.~Stachniss, ``Rangenet ++: Fast and
  accurate lidar semantic segmentation,'' in {\em 2019 IEEE/RSJ International
  Conf. on Intelligent Robots and Systems (IROS)}, pp.~4213--4220, 2019.

\bibitem{guo2018review}
Y.~Guo, Y.~Liu, T.~Georgiou, and M.~S. Lew, ``A review of semantic segmentation
  using deep neural networks,'' {\em International journal of multimedia
  information retrieval}, vol.~7, no.~2, pp.~87--93, 2018.

\bibitem{fernandes2021point}
D.~Fernandes, A.~Silva, R.~N{\'e}voa, C.~Sim{\~o}es, D.~Gonzalez, M.~Guevara,
  P.~Novais, J.~Monteiro, and P.~Melo-Pinto, ``Point-cloud based 3d object
  detection and classification methods for self-driving applications: A survey
  and taxonomy,'' {\em Information Fusion}, vol.~68, pp.~161--191, 2021.

\bibitem{s3dis}
I.~Armeni, O.~Sener, A.~R. Zamir, H.~Jiang, I.~Brilakis, M.~Fischer, and
  S.~Savarese, ``3d semantic parsing of large-scale indoor spaces,'' in {\em
  2016 IEEE Conference on Computer Vision and Pattern Recognition (CVPR)},
  pp.~1534--1543, 2016.

\bibitem{semantic3d}
T.~Hackel, N.~Savinov, L.~Ladicky, J.~D. Wegner, K.~Schindler, and
  M.~Pollefeys, ``Semantic3d.net: {A} new large-scale point cloud
  classification benchmark,'' {\em CoRR}, vol.~abs/1704.03847, 2017.

\bibitem{behley2019semantickitti}
J.~Behley, M.~Garbade, A.~Milioto, J.~Quenzel, S.~Behnke, C.~Stachniss, and
  J.~Gall, ``Semantickitti: A dataset for semantic scene understanding of lidar
  sequences,'' in {\em Proceedings of the IEEE/CVF International Conference on
  Computer Vision}, pp.~9297--9307, 2019.

\bibitem{paris2017}
X.~Roynard, J.~Deschaud, and F.~Goulette, ``Paris-lille-3d: a large and
  high-quality ground truth urban point cloud dataset for automatic
  segmentation and classification,'' {\em CoRR}, vol.~abs/1712.00032, 2017.

\bibitem{Dales}
N.~M. Varney, V.~K. Asari, and Q.~Graehling, ``{DALES:} {A} large-scale aerial
  lidar data set for semantic segmentation,'' {\em CoRR}, vol.~abs/2004.11985,
  2020.

\bibitem{Semanticposs}
Y.~Pan, B.~Gao, J.~Mei, S.~Geng, C.~Li, and H.~Zhao, ``Semanticposs: {A} point
  cloud dataset with large quantity of dynamic instances,'' {\em CoRR},
  vol.~abs/2002.09147, 2020.

\bibitem{kitti360}
Y.~Liao, J.~Xie, and A.~Geiger, ``{KITTI}-360: A novel dataset and benchmarks
  for urban scene understanding in 2d and 3d,'' {\em arXiv preprint
  arXiv:2109.13410}, 2021.

\bibitem{mohammadi2020mixed}
F.~Mohammadi~Amin, M.~Rezayati, H.~W. van~de Venn, and H.~Karimpour, ``A
  mixed-perception approach for safe human--robot collaboration in industrial
  automation,'' {\em Sensors}, vol.~20, no.~21, p.~6347, 2020.

\bibitem{Jiang1987ACA}
B.~Jiang and C.~Gainer, ``A cause-and-effect analysis of robot accidents,''
  {\em Journal of Occupational Accidents}, vol.~9, pp.~27--45, 1987.

\bibitem{9196924}
M.~El-Shamouty, X.~Wu, S.~Yang, M.~Albus, and M.~F. Huber, ``Towards safe
  human-robot collaboration using deep reinforcement learning,'' in {\em 2020
  IEEE International Conference on Robotics and Automation (ICRA)},
  pp.~4899--4905, 2020.

\bibitem{Hungry}
B.~Gao, Y.~Pan, C.~Li, S.~Geng, and H.~Zhao, ``Are we hungry for 3d lidar data
  for semantic segmentation?,'' {\em CoRR}, vol.~abs/2006.04307, 2020.

\bibitem{thomas2019kpconv}
H.~Thomas, C.~R. Qi, J.-E. Deschaud, B.~Marcotegui, F.~Goulette, and L.~J.
  Guibas, ``Kpconv: Flexible and deformable convolution for point clouds,'' in
  {\em Proceedings of the IEEE/CVF International Conference on Computer
  Vision}, pp.~6411--6420, 2019.

\bibitem{hu2020randla}
Q.~Hu, B.~Yang, L.~Xie, S.~Rosa, Y.~Guo, Z.~Wang, N.~Trigoni, and A.~Markham,
  ``Randla-net: Efficient semantic segmentation of large-scale point clouds,''
  in {\em Proceedings of the IEEE/CVF Conference on Computer Vision and Pattern
  Recognition}, pp.~11108--11117, 2020.

\end{thebibliography}

\end{document}